\documentclass[10pt,twocolumn,letterpaper]{article}

\usepackage{wacv}
\usepackage{times}
\usepackage{epsfig}
\usepackage{graphicx}
\usepackage{amsmath}
\usepackage{amssymb}
\usepackage[export]{adjustbox}

\newcommand{\figref}[1]{Fig.~\ref{#1}}
\newcommand{\tblref}[1]{Table~\ref{#1}}
\newcommand{\sref}[1]{Section~\ref{#1}}

\usepackage{graphicx}
\makeatletter
\@namedef{ver@everyshi.sty}{}
\makeatother
\usepackage{tikz}

\usepackage[pagebackref=true,breaklinks=true,colorlinks,bookmarks=false]{hyperref}

\wacvfinalcopy % *** Uncomment this line for the final submission

\def\wacvPaperID{} % *** Enter the wacv Paper ID here

\begin{document}

\title{Synthesizing human-like sketches from natural images\\ using a conditional convolutional decoder}

\author{Moritz Kampelm\"uhler, Axel Pinz\\
Graz University of Technology\\
Graz, Austria\\
{\tt\small kampelmuehler@tugraz.at}
}

\maketitle
%\thispagestyle{empty}

%-------------------------------------------------------------------------
%%%%%%%%% ABSTRACT
\begin{abstract}
Humans are able to precisely communicate diverse concepts by employing sketches, a highly reduced and abstract shape based representation of visual content. We propose, for the first time, a fully convolutional end-to-end architecture that is able to synthesize human-like sketches of objects in natural images with potentially cluttered background. To enable an architecture to learn this highly abstract mapping, we employ the following key components: (1) a fully convolutional encoder-decoder structure, (2) a perceptual similarity loss function operating in an abstract feature space and (3) conditioning of the decoder on the label of the object that shall be sketched. Given the combination of these architectural concepts, we can train our structure in an end-to-end supervised fashion on a collection of sketch-image pairs. The generated sketches of our architecture can be classified with 85.6\% Top-5 accuracy and we verify their visual quality via a user study. We find that deep features as a perceptual similarity metric enable image translation with large domain gaps and our findings further show that convolutional neural networks trained on image classification tasks implicitly learn to encode shape information. Code is available under \url{https://github.com/kampelmuehler/synthesizing_human_like_sketches}

\end{abstract}

%-------------------------------------------------------------------------
%%%%%%%%% INTRODUCTION
\section{Introduction}

A human sketch of an object contained in a natural image constitutes a highly abstract representation. For a particular input image, the sketch will vary significantly, depending on the person who actually draws it, but also on drawing skills, context contained in the remainder of the image, and possibly even on the current mood of the artist. We aim to find a transformation from natural RGB images to human-like sketches of objects contained in these images. From a trivial point of view, it would suffice to extract edge maps and to constrain them to the object regions. However, this is not what we consider to be a `human-like' sketch. Definitions of what human-like means in the context of sketches are found in works on sketch based image retrieval and sketch recognition like \cite{eitz2012} and \cite{sangkloy2016}. In general, human sketches are comprised of salient inner and outer contours that reflect the basic shape of objects, including a level of abstraction that might be far from calculated edge images. Despite this abstraction, humans in most cases are able to recognize sketches very effectively. Imagine for example the sketch of a rabbit. Characteristic features that come to mind are the long ears, bushy tail and long hind legs. In actual images of rabbits these features might not be very prominent, if at all present. Still, human sketchers will most certainly include them in sketches to convey the concept of a rabbit. Apparently, for humans it is more important to convey the basic notion of an object via a sketch, rather than an exact representation of a specific instance like in a constrained edge image. Our goal is thus not to synthesize artistic sketches of objects, as it is for example done in the non-photorealistic rendering community \cite{winnemoller2012}, but rather to find out whether convolutional neural networks (CNNs) are able to learn abstract shape representations.

\begin{figure}[t]
\begin{center}
   \includegraphics[width=0.98\linewidth]{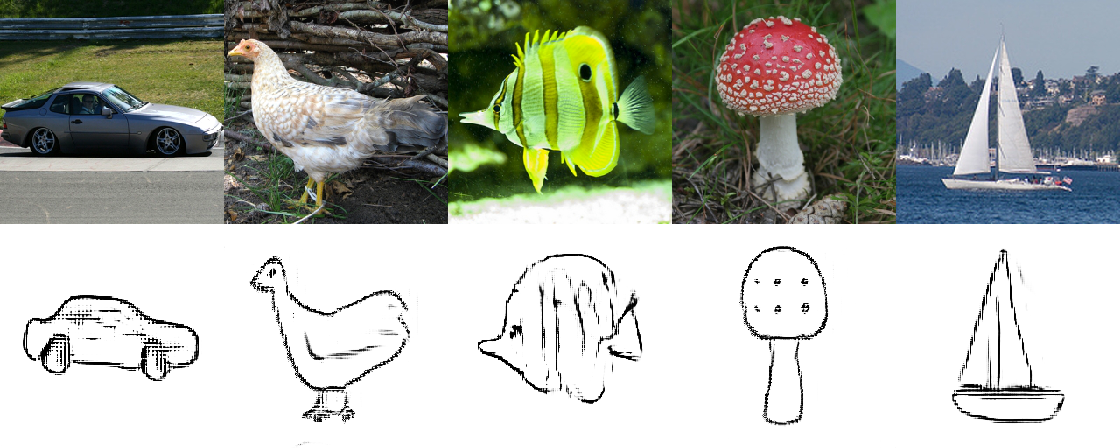}
\end{center}
\vspace{-12pt}
   \caption{Some examples of sketches generated from images using our proposed method. Our fully convolutional encoder-decoder architecture is able to generate abstract human-like sketches from natural images with cluttered background without the need for any pre- or post-processing.}
\label{fig:example_predictions}
\vspace{-6pt}
\end{figure}

 Recent work by Geirhos \etal \cite{geirhos2019} suggests that CNNs trained on image classification are biased towards texture. Their findings are directly opposed to the work of Kubilius \etal \cite{kubilius2016} who show that such architectures arrive at useful implicit shape representations that are somewhat comparable to human abstraction capabilities. While we believe that shape as such is indeed an underrepresented entity in recent computer vision work, especially in recognition, we suggest that implicit shape representations learned by CNNs facilitate more complex shape related tasks. We employ a fully convolutional encoder-decoder structure to accomplish a mapping from image space to sketch space, and thereby present the first viable approach to produce human-like sketches from natural, potentially cluttered input images (see \figref{fig:example_predictions}). Our findings further confirm what Kubilius \etal \cite{kubilius2016} suggest, by showing that strongly shape related tasks can indeed be solved on top of the feature representation of CNNs trained on image classification tasks.
Our approach may be useful for augmenting sketch-based retrieval databases, creative applications, or object boundary level computer vision algorithms, among others.

%------------------------------------------------------------------------
%%%%%%%%% RELATED WORK
\section{Related Work}

%--------------------
% Sketches
\paragraph{Sketch-based retrieval and sketch recognition.}

The earliest traces of sketch-based image retrieval were running under the name of QVE (Query by Visual Example). Hirata and Kato \cite{hirata1992} came up with the idea of reducing images to refined edge maps they call \textit{pictorial indices} which can then be matched with more abstract representations like sketches in a correlational manner. A first collection of objects sketched by humans was made by Eitz \etal \cite{eitz2012}, which comprises 250 object categories. In addition to collecting this dataset, they also came up with a sketch representation based on histograms of visual words that can be used for sketch recognition and sketch-based retrieval. Sangkloy \etal \cite{sangkloy2016} later were the first to source a large scale collection of image-sketch pairs dubbed the \textit{Sketchy Database}, consisting of 125 object classes. Their goal was to provide a solid benchmark for sketch-based retrieval algorithms and to also test their novel deep learning based retrieval method. In their top performing approach they first train independent classifiers for sketches as well as images and later introduce a triplet loss to enforce minimum distance between the individual embeddings. We use the \textit{Sketchy Database} as dataset for training and evaluation. Sangkloy \etal \cite{sangkloy2016} extract features using the \textit{GoogLeNet} architecture \cite{szegedy2015}, which is designed for image recognition. Yu \etal \cite{yu2017} propose a specialized CNN for sketch recognition with the key components being larger filters in the first layer, no normalization layers and larger pooling kernels. While achieving competitive results, they unfortunately do not compare their recognition scores with those obtained with \textit{GoogLeNet}. In our work, we use a VGG16 CNN architecture for image recognition \cite{simonyan2015}, because it is well established and well suited for a symmetric hierarchical encoder-decoder path.

%--------------------
% Generative Modeling
\vspace{-8pt}
\paragraph{Generative modeling.}
Since the creation of Generative Adversarial Networks \cite{goodfellow2014} (GANs) interest in generative modeling is rapidly rising within the computer vision community and regular advances are made in various domains such as  face image generation \cite{karras2018}, image inpainting \cite{yu2018} and video generation \cite{saito2017}. Yet, not only GANs have contributed to the state of the art, but also Encoder-Decoder architectures have seen a rise in popularity for diverse applications. Examples include monocular depth estimation \cite{godard2017, zhou2017}, where dense depth maps are generated from monocular camera images, and optical flow estimation \cite{ilg2017}, where pixel wise displacements between two images are tracked. For the task at hand, we decide to rely on an Encoder-Decoder style architecture, since such architectures are more accessible regarding training and interpretability.

The problem opposite to image-to-sketch synthesis is lately frequently tackled in research. Popular approaches that model sketch based image synthesis are the GAN based architectures \textit{pix2pix} \cite{isola2017} and \textit{SketchyGAN} \cite{chen2018}. While \textit{pix2pix} is proposed as a generic architecture for image translation tasks, \textit{SketchyGAN} specifically aims to apply texture to given outlines, which is somewhat related to image inpainting \cite{yu2018}. Interestingly, Chen and Hays \cite{chen2018} begin training on pairs of edge maps and images and gradually fade towards sketch-image pairs. This shows the inherent difficulty of immediately mapping from the highly abstract sketch space to image space, that even established methods like \textit{pix2pix} can not readily achieve.

To the best of our knowledge, the only published line of work in multi-class sketch generation is \textit{Sketch-RNN} \cite{ha2018}, which is a recurrent architecture operating on a vectorized brush stroke level. Hence, it is fully decoupled from the image domain and also does not aim at establishing correspondences between images and sketches. More recently, Muhammad \etal \cite{muhammad2018} have proposed a method for photo-to-sketch synthesis, which operates on top of edge images. They train an agent in a reinforcement learning setting to learn which elements of an edge image can be removed without critically diminishing its recognizability and present results on images of shoes with homogeneous white background. The crucial differences to our goals are that we want to (1) abstain from edge maps, (2) be able to handle multiple classes and (3) infer sketches from natural images with potentially cluttered background. Furthermore, Muhammad \etal \cite{muhammad2018} claim that deep encoder-decoder models are incapable of handling the large domain gap and misalignment between sketches and images, which we show is not necessarily the case.

\begin{figure*}[t]
\begin{center}
\includegraphics[width=\textwidth]{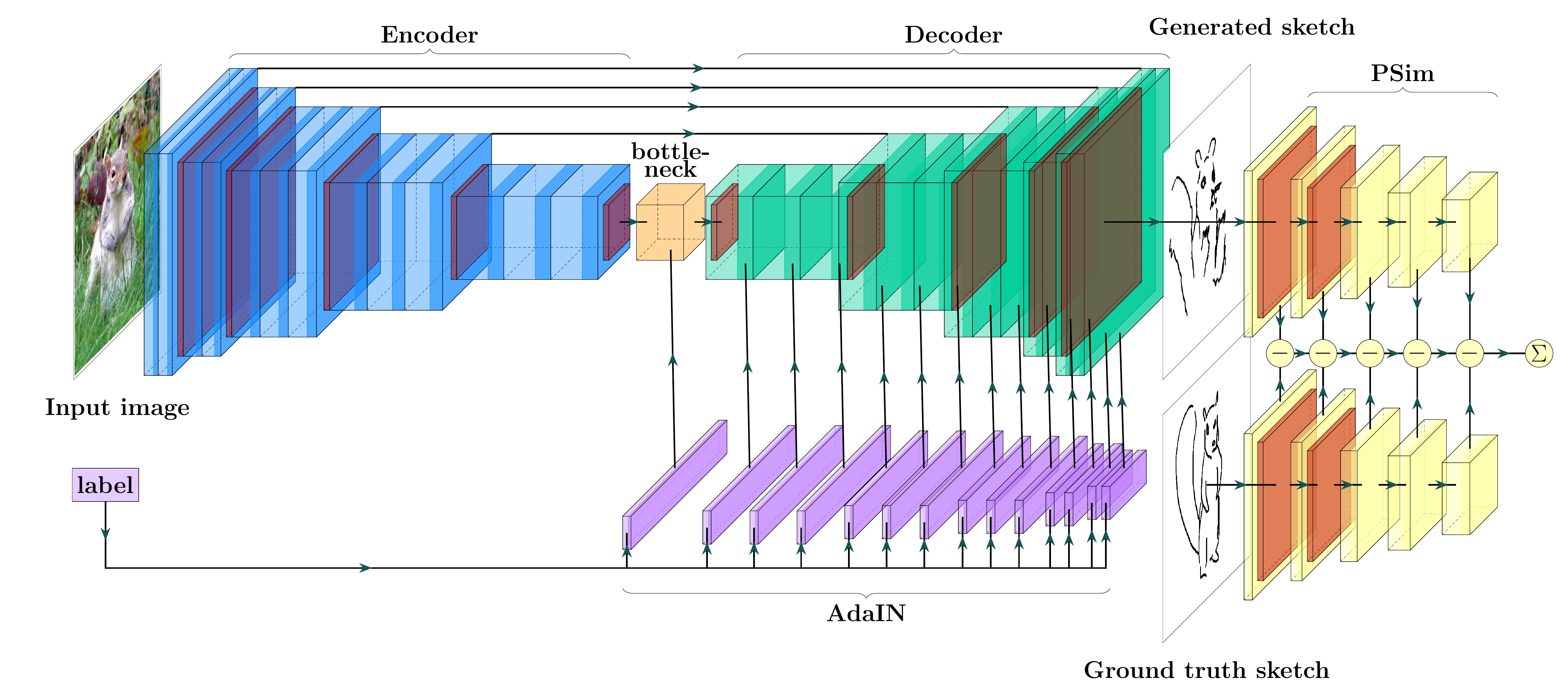}
\end{center}
\vspace{-12pt}
   \caption{An overview of our architecture. The input image is transformed into feature space (denoted as bottleneck) and then transformed into a sketch via transpose convolutions. Batch normalization in the decoder is replaced with Adaptive Instance Normalization to condition on a certain label. The arrows from the encoder to the decoder layers denote skip connections, such that the activations of the respective layers in the encoder are concatenated with the inputs of the corresponding layers of the decoder. The right part of the figure shows two AlexNet trunks that extract features from generated and ground truth sketch. The loss for the decoder is then defined as the cosine similarity of the features of each layer. Only the decoder and AdaIN embedding parameters are updated during training.}
\label{fig:architecture}
\vspace{-6pt}
\end{figure*}

%--------------------
% Perceptual similarity metrics
\vspace{-8pt}
\paragraph{Perceptual similarity metrics.}
Careful choice and construction of adequate loss functions is crucial for achieving visually pleasing results in generative models. Larsen \etal \cite{larsen2016} show that pixel-wise distances like \textit{L2} tend to yield blurry results in encoder-decoder networks, when used as sole loss function. Visual appeal implies that the perceptual similarity of generated samples with real samples needs to be maximized. A more traditional image quality metric, \textit{Structural Similarity} (SSIM), was proposed by Wang \etal \cite{wang2004}. The basic idea behind SSIM lies in the comparison of image patches regarding their luminance, contrast and structure, which are represented by mean, variance and cross-correlation respectively. They also proposed an extension,  MSSSIM, which processes each patch at multiple scales. These metrics have also been used successfully in deep encoder-decoder structures later on \cite{ridgeway2017}.

More recently, interest has surged in using deep similarity metrics for image generation and reconstruction. The idea of Dosovitskiy and Brox \cite{dosovitskiy2016} lies in not only calculating distances in image space, but also in feature space of neural networks. The basic notion of this technique is that perceptually similar images shall map to feature vectors with a small mutual distance given the same network parameters. In \cite{dosovitskiy2016} only single later layers of the CNNs are used to compute distances, which can lead to ambiguous results due to many inputs mapping to proximal locations in the more abstract latent space. In many cases this ambiguity is undesirable and can be relieved by calculating distances between feature vectors of the individual layers in the CNN \cite{zhang2018}. We find that the cosine similarity between layer activations proposed by Zhang \etal \cite{zhang2018} serves the task of image generation especially well and that we achieve best results relying on this metric as reconstruction loss.

%------------------------------------------------------------------------
%%%%%%%%% METHOD SECTION

\section{Method}

Our method is comprised of several key components depicted in \figref{fig:architecture}. The architecture is built around a convolutional decoder. This decoder reconstructs a sketch from a latent representation of an input image obtained via an encoder that is a mirrored version of the decoder. Additionally, the individual layers of the decoder can be conditioned on the label of the input image. Lastly, a loss function operating on abstracted features of a CNN enables training through backpropagation. Only the weights of the decoder and label conditioning are updated during training. The remainder of this section goes into more detail about the individual parts.

\subsection{Network Architecture}\label{sec:architecture}

The basic building block of our architecture is a fully convolutional encoder-decoder structure. Its working principle is to first transform a 224x224x3 input image in RGB space into an abstract 7x7x512 dimensional feature space referred to as the bottleneck and then reconstructing a 224x224x1 sketch from the bottleneck representation. It is made up of the convolutional stack of a VGG16 \cite{simonyan2015} network with Batch Normalization as an encoder and a decoder that is a mirrored version of the encoder with the pooling layers replaced by bilinear upsampling. We found that using bilinear upsampling as opposed to Max Unpooling \cite{noh2015} fosters the training process as it allows for smoother gradient flow in the decoder. The encoder weights are pretrained on ImageNet \cite{russakovsky2015} and frozen, meaning that they are not updated during training. The decoder is randomly initialized using \textit{Kaiming Uniform} \cite{he2015} initialization, with the same random seed for all experiments. Additionally, we introduce skip connections, as proposed by Ronneberger \etal \cite{ronneberger2015}, from the encoder activations to the decoder inputs. We concatenate the outputs of the convolutional blocks in the encoder before each instance of Max Pooling with the corresponding blocks in the decoder after each upsampling operation. The resulting four shortcuts (see also \figref{fig:architecture}) enable the decoder to also incorporate lower level feature representations into the reconstruction. To accommodate the increased number of input channels from the skip connections in the decoder, we employ two different strategies: (1) we reduce the number of input channels before the upconvolution layers using 1x1 convolutions, and (2) we increase the number of input channels of the upconvolution layers to accommodate the increased number of input features. The first method introduces one million additional parameters, while the latter adds just over three million parameters.

Within the decoder all activations are normalized using Adaptive Instance Normalization (AdaIN) \cite{huang2017} with class embeddings instead of standard Batch Normalization. AdaIN is defined as 

\begin{equation}
\label{eq:AdaIN}
\textrm{AdaIN}(x, x_t)= \sigma(x_t)\left(\frac{x-\mu(x)}{\sigma(x)}\right)+\mu(x_t)
\end{equation}

and originally used for image style transfer, where the style (\ie the instance statistics $\mu(x),~\sigma(x)$) of an image $x$ is removed and replaced with the style of a reference image  ($\mu(x_t),~\sigma(x_t)$). In our implementation we use AdaIN to condition the decoder on the class of the desired sketch, which induces a `shape prior' within the decoder. We achieve this by leveraging embedding layers, which are layers primarily used in Natural Language Processing \cite{mikolov2013}. They are used to embed individual words of a vocabulary into a vectorized feature space by learning representations that shall ideally cluster similar words in feature space. In our case we learn an embedding for the mean and standard deviation of each class and each featuremap. These embeddings then allow us to apply learned class related statistics to each feature representation in the decoder by replacing Batch Normalization in the decoder with AdaIN \eqref{eq:AdaIN}. In contrast to appending a label vector to the input of the decoder, like it is often implemented in GANs \cite{mirzaO14}, this method enables us to more explicitly inject label information at various abstraction levels in the decoder and we  find that it yields superior performance to the former method. 

Our method refrains from using a pixel wise loss for reconstruction. Instead we use a perceptual similarity metric as introduced in \cite{zhang2018}, which builds on top of an AlexNet \cite{krizhevsky2012} convolutional stack. We first train a full AlexNet CNN as a sketch classifier for the Sketchy Database (see \sref{sec:sketchydb}) and then remove the fully connected classification head to then use the learned weights of the convolutional layers for the similarity metric. For the generated and target sketches the loss is then the cosine similarity between the activation vectors of the individual convolutional layers:

\begin{equation}
d(x,x_t) = \sum_l \dfrac{1}{H_l W_l} \sum_{h,w} ||  \hat{y}_{hw}^l - \hat{y}_{t,hw}^l  ||_2^2,
\label{eq:dist}
\end{equation}

where $x$ and $x_t$ are the inputs, $l$ denotes the layer, $H_l$ and $W_l$ denote the height and width dimension of the activations of layer $l$ and $\hat{y}_{hw}^l$ and $\hat{y}_{t,hw}^l$ are the channels of the activations of layer $l$ for $x$ and $x_t$ respectively.
We choose this perceptual similarity loss as a superior alternative to pixel wise distances, which introduce numerous difficulties to the training of generative models \cite{dosovitskiy2016}. Imagine, for example, a checkerboard pattern: If shifted by just one box in either direction, a pixel wise distance changes from minimum to maximum. This is especially apparent in the sketch domain, in which quasi binary images with thin strokes are compared and small displacements that do not influence perceptual quality might result in irrationally large pixel-wise distances. The perceptual similarity metric we employ in this work is, through the use of CNNs, invariant to such small shifts.

\subsection{Network Training}

While the final architecture can be trained in an end-to-end supervised manner, the convolutional stack of the loss function requires pre-training before being plugged into the final architecture. We train the AlexNet model used for the loss as a classifier for sketches first. For this purpose, we use an augmented version of all the sketches in the Sketchy Database and the dedicated labels. Our augmentation regime consists of: horizontal flips of all sketches, random rotation of $[-10 ^o, 10 ^o]$, random horizontal and vertical translation of $[-18\text{px}, 18\text{px}]$, random scale of $[-10\%, 10\%]$ and shear of $[-10 ^o, 10 ^o]$. We split the dataset into training and test set 80/20 which results in 60218 training and 15055 test sketches and then train for 130 epochs with the Adam \cite{kingma2014} optimizer and Dropout \cite{srivastava2014} of 50\%. After training, the classifier achieves a Top-1 accuracy of 84 \% on the test split which is sufficiently accurate for the use as a feature extractor. For use as the loss function of the entire architecture, we discard the fully connected layers of the AlexNet classifier and only use the convolutional stack.

The use of AlexNet for our perceptual similarity loss is not arbitrary. In terms of predictive power, AlexNet is no longer state of the art and has long been superseded by the likes of ResNets \cite{he2015}, or Inception (GoogLeNet) networks \cite{szegedy2015} and successors. Nonetheless, the feature representation of AlexNet appears to have certain benefits for the use as a feature loss. When experimenting with ResNets of various depths as well as Inception-v3 \cite{szegedy2016}, we found that the outputs are somewhat similar to those of pixel level losses and the resulting architectures tend to overfit, rather than generalize. We conclude that this behavior is to be expected given the architectures of those networks. In AlexNet the convolutional layers are truly stacked representations at fixed scales and fixed input strides and thus are clear \textit{`features of features'}. ResNets, on the other hand, rely on residual connections for training very deep networks and these residual connections also provide access to lower level features at arbitrary points in the network. Similarly, in Inception networks each layer operates on its input at various scales at once. Both of the more powerful architectures thus provide access to lower level features also at higher levels and consequently the objective that is optimized in the perceptual loss appears to emphasize those low level representations.

Once the feature extractor for the loss function is pre-trained, we can add it to the full architecture to train it end-to-end with full supervision. In this end-to-end regime we use RGB images as input and the corresponding sketches as the targets for supervision. We augment all the input images with a random color jitter of 10\% in brightness, contrast and saturation to prevent the network from remembering the mappings from RGB to sketch. As optimizer we again use Adam since it showed to achieve faster convergence and added training stability compared to stochastic gradient descent with momentum and learning rate scheduling. All activation functions are set to ReLu except for the output layer of the decoder which uses a sigmoid activation to squash the output sketch into the range of $[0,1]$ of the target sketch. The gradient flows from the output of the perceptual loss back to the bottleneck, but the only weights that are updated according to the local partial derivatives are those of the upconvolutional layers and AdaIN in the decoder. Additional training of the encoder does not increase the performance of our method, but rather decreases it. Since the dataset is very limited in size, the increased number of trainable parameters introduced by training the encoder is not feasible in training and leads to overfitting. Training the encoder yields the best accuracy on the training split but can not outperform a method with frozen encoder weights on the test split. We consider training to converge once the accuracy on the test set (see \sref{sec:sketchydb}) saturates which on average takes 300 epochs. Notably, for each input image several output sketches exist (see \sref{sec:sketchydb} and \figref{fig:example_dataset}) which means that for every given input there is no single `right answer'. While this confusion breaks training with pixel-wise loss functions, the perceptual similarity loss is able to handle and even benefit from multiple targets without issues.

%------------------------------------------------------------------------
%%%%%%%%% EXPERIMENTS

\section{Experiments}\label{sec:experiments}

This section serves to provide extensive experimental evaluation of our proposed approach with quantitative as well as qualitative results. We seek to not only identify the parts that work well, but also those that fail or require improvement. First of all we provide an overview of the dataset used for our experiments.

%--------------------
% DATASET
\subsection{The Sketchy Database}\label{sec:sketchydb}

\begin{figure}[ht]
\vspace{-12pt}
\begin{center}
   \includegraphics[width=0.85\linewidth]{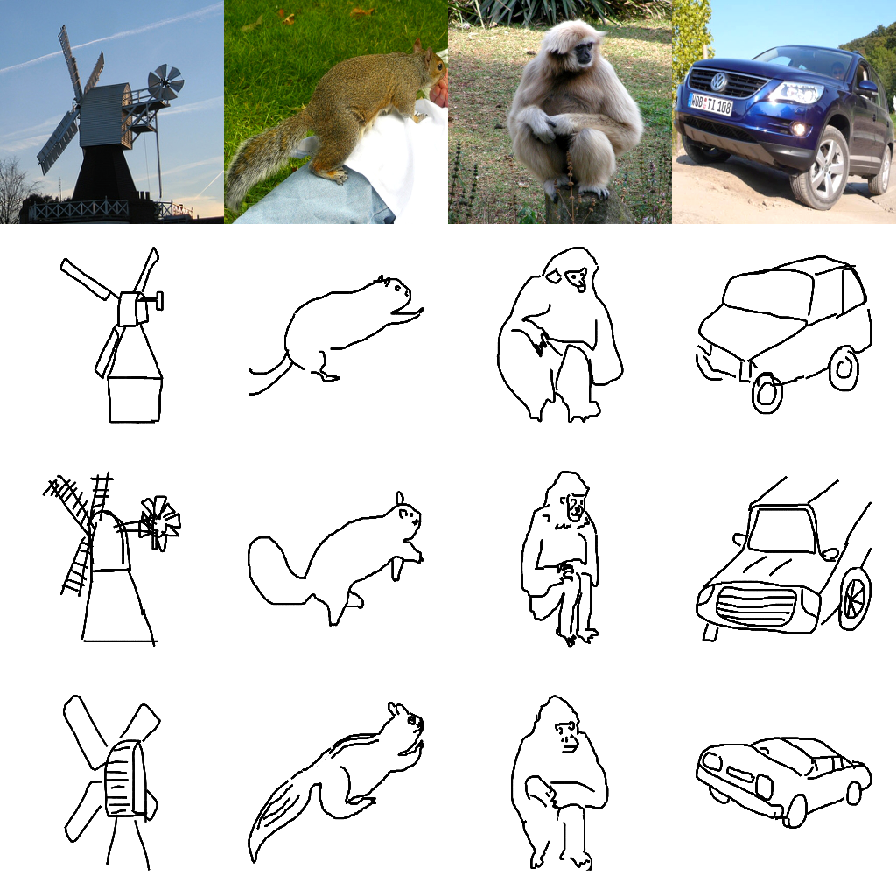}
\end{center}
\vspace{-12pt}
   \caption{Example images from the Sketchy Database with sketches drawn by different individuals. Note the apparent differences between different sketches for the same input image as well as the levels of abstraction.}
\label{fig:example_dataset}
\vspace{-6pt}
\end{figure}

For this work we use the Sketchy Database \cite{sangkloy2016} dataset which is a large collection of sketch-image pairs. It consists of 125 distinct classes with natural images taken from the ImageNet \cite{russakovsky2015} dataset. For each image, several sketches are collected from different sketch artists, with an average of 6 and a minimum of 5 sketches per image. Examples are given in \figref{fig:example_dataset}. Notably, the sketchers are not allowed to trace or copy the sketches, they may only look at the images for two seconds at a time. This ensures a richer dataset with abstracted sketch images. For our work, we use images cropped to the annotated bounding box, sketches that are centered and scaled to the bounding box, and omit sketches marked as erroneous, ambiguous, containing context, or showing wrong pose. We employ a 90/10 train/test split of the dataset leaving us with 11193 images and 57811 image-sketch pairs for training and 1274 images for testing. Unfortunately, for this task it is necessary to drop a fair amount of sketch data, since at test time we only need the images and in the split of the data it is important to split on images rather than sketch-image pairs in order to guarantee unseen images in the test set.

\subsection{Evaluation metric}\label{sec:eval_metric}

\begin{figure*}[t]
\begin{center}
  \includegraphics[width=0.98\linewidth]{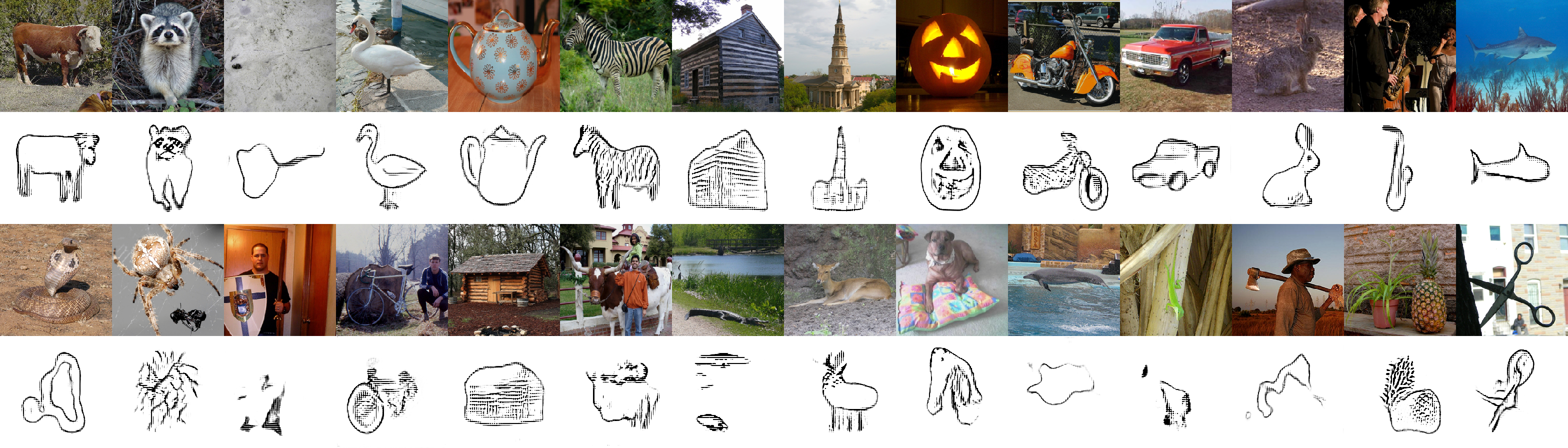}
\end{center}
\vspace{-12pt}
  \caption{Positive (top) and negative (bottom) reconstruction samples from our approach. Note that no post processing is applied and thus the output exhibits some high frequency artifacts. Our approach works well, if the object of concern is not occluded, sufficiently fills the frame and is not displayed with similar objects adjacent to it. Failures occur especially if the pose of the object is rare (deer in the middle), if the object is occluded (bicycle on the left), does not appear in reasonable size (crocodile in the middle), or is confused with similar objects nearby (pineapple on the right).}
\label{fig:example_good_bad}
\vspace{-6pt}
\end{figure*}

\begin{table}[t]
\begin{tabular}{|c||c|c|c|}
\hline
method & Top-1 & Top-5 & \#params \\ \hline \hline
chance  &  0.8\%     &   3.9\%    &   -           \\ \hline
HED  &  0.4\%     &   3.2\%    &   14.7M          \\ \hline
MSE  &  1.3\%     &   4.9\%    &   \textbf{17.0M}           \\ \hline \hline
PSim    &  37.9\%     &   60.2\%    &      \textbf{17.0M}        \\ \hline
PSim+flip  &  47.1\%     &   69.2\%    &       \textbf{17.0M}       \\ \hline
PSim+flip+AdaIN &   61.4\%    &    79.9\%   &       18.2M       \\ \hline
PSim+flip+AdaIN+skip1 &   62.3\%    &    80.7\%   &    19.2M          \\ \hline
PSim+flip+AdaIN+skip  &   \textbf{66.6\%}   &  \textbf{85.6\%}     &  21.3M            \\ \hline\hline
ground truth  &   91\%  &  98\%     &  -            \\ \hline
\end{tabular}
\vspace{3pt}
\caption{\label{tab:results}Classification results of generated sketches for different parts of the full architecture. The first block denotes baselines: random guess, HED \cite{xie2015} edges and our architecture trained with MSE loss. The methods within the second block use the perceptual similarity loss. flip denotes random horizontal flipping of sketches and images, AdaIN denotes our proposed conditioning on labels, skip1 denotes connections from the encoder to the decoder using 1x1 convolutions to preserve feature map channel dimensions and skip denotes such connections but with the number of input channels of subsequent upconvolutional layers increased. The ground truth accuracy is the accuracy of the classifier on sketches from the dataset.}
\vspace{-6pt}
\end{table}

To objectively quantify the performance of our methods, we choose to train another classifier on all of the sketches from the whole dataset. We deliberately choose a 34 layer ResNet architecture here, since in experiments with various network depths the 34 layer structure yielded the best compromise between predictive power and generalization capability. Also, we refrain from recycling the AlexNet classifier from the feature loss to ensure that the models used for training and evaluation are independent.
After 130 epochs of training on an 80/20 train/test split using Adam as optimizer, we obtain a classifier with a Top-1 accuracy of 91\% and a Top-5 accuracy of 98\%. Using this metric ensures that the sketches that are synthesized not only look similar to sketches from the Sketchy database but also conform to the class of the input image. Hence, we rank our experiments by the achieved ResNet 34 Top-1 and Top-5 accuracy.

This metric can not necessarily judge the conformity of the output sketch to the input image. Due to the highly abstract nature of human sketches this is also a score that is inherently hard to establish. Have a look at the fourth column of \figref{fig:example_dataset}: All the sketches can be recognized as a car, also the relative pose of the sketches is to some extent concordant with the image. However, from the sketches alone one most probably will not be able to come to the conclusion that they depict an SUV, let alone a VW Touareg. So the truth of the matter is that there does not necessarily exist a mapping from image to sketch or vice versa, but rather the image serves as a rough guideline for the sketcher. This is exactly what exemplifies the difficulty as well as the appeal of this task.

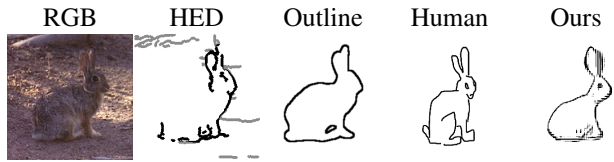
\begin{figure}
\centering
\begin{tikzpicture}[picture format/.style={inner sep=0pt,}]

  \node[picture format] (A1)               {\includegraphics[width=0.2\linewidth]{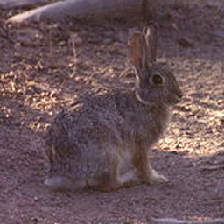}};
  
  \node[picture format,anchor=north west] (A2) at (A1.north east) {\includegraphics[width=0.2\linewidth]{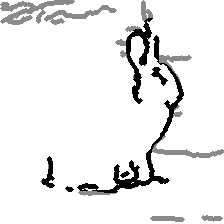}};

  \node[picture format,anchor=north west] (A3) at (A2.north east) {\includegraphics[width=0.2\linewidth]{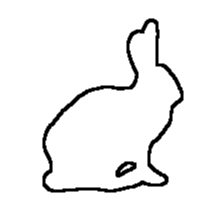}};
  
  \node[picture format,anchor=north west] (A4) at (A3.north east)      {\includegraphics[width=0.2\linewidth]{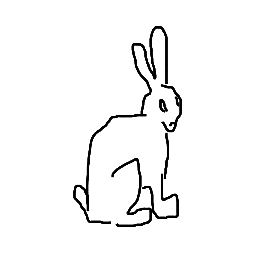}};

  \node[picture format,anchor=north west] (A5) at (A4.north east) {\includegraphics[width=0.2\linewidth]{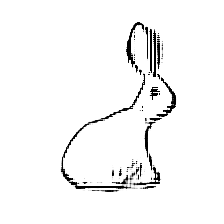}};

  %% Captions

  \node[anchor=base, shift={(0,0.1)}] (C1) at (A1.north) {RGB};
  \node[anchor=base, shift={(0,0.1)}] (C2) at (A2.north) {HED};
  \node[anchor=base, shift={(0,0.1)}] (C3) at (A3.north) {Outline};
  \node[anchor=base, shift={(0,0.1)}] (C4) at (A4.north) {Human};
  \node[anchor=base, shift={(0,0.1)}] (C5) at (A5.north) {Ours};

\end{tikzpicture}
\caption{A comparison of our approach with a (manually masked) HED edge map of the input image, the outline of the segmentation mask and a ground truth sketch. The masked edge extraction method fails to provide a meaningful representation. The outline of the segmentation mask is sufficient, but does not conform to our notion of a human-like sketch and misses inner contours, in this case specifically the eye. Our method yields a more abstract sketch of the rabbit in the correct pose comparable to the human ground truth sketch.}
\label{fig:edges_mask}
\vspace{-20pt}
\end{figure}

\begin{figure}[t]
\begin{center}
   \includegraphics[width=0.98\linewidth]{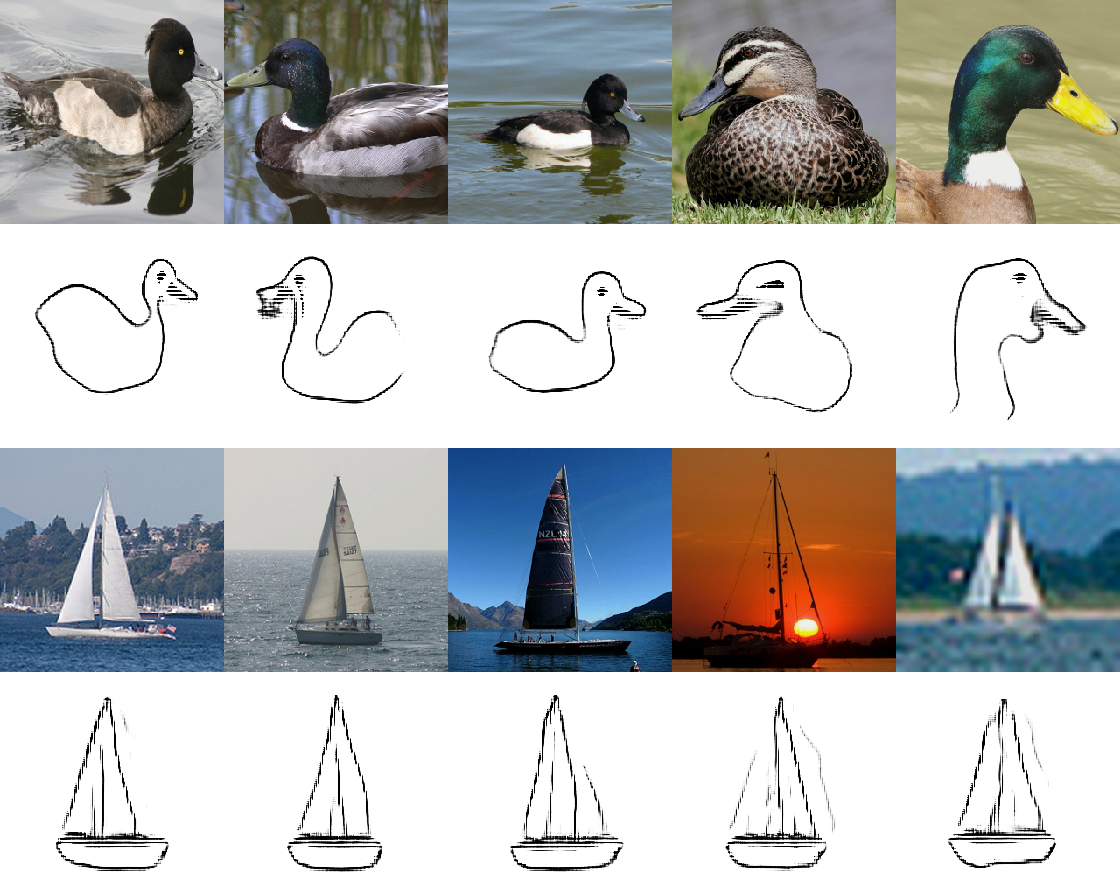}
\end{center}
\vspace{-12pt}
   \caption{Several generated samples from the two distinct classes duck and sailboat. The dataset includes ducks in various poses, which enables the decoder to generate samples with higher intra-class variance. For sailboats, the data is limited to mostly profile views, such that the intra-class variance is low, which looks like signs of a mode collapse, but rather is rooted in the training data.}
\label{fig:example_mode}
\vspace{-14pt}
\end{figure}

%-----------------------
%%%%%%%%% Ablation Study

\subsection{Ablation study}\label{sec:ablation}

Our final approach consists of several key components contributing to its performance according to the evaluation metric described in \sref{sec:eval_metric}. In order to judge the relative performance improvements we provide an ablation study in which we train architectures with key components of our full architecture removed and quantitatively evaluate each of the models we thus obtain. We train all of the models for 300 epochs on an identical split of the dataset. 

\tblref{tab:results} shows the evaluation results for the different levels of our approach. Clearly, the baselines of using HED edge extraction \cite{xie2015} or training our architecture with an MSE loss yield no usable results for sketch generation. They are virtually on par with the respective chance levels. Replacing the loss function with the perceptual similarity metric introduced in \sref{sec:architecture} and introducing horizontal flipping as data augmentation pushes the Top-1 accuracy by 45.8\% and the Top-5 accuracy by 64.3\%. Conditioning the decoder on the label of the input image using adaptive instance normalization gives another increase of 14.3\% and 10.7\% in Top-1 and Top-5 accuracy respectively. Notably, the addition of the conditioning also increases the number of parameters by 6.6\% or 1.2 million. At the cost of another increase in the number of parameters of 14.6\% or 3.1 million we can achieve a further increase in Top-1 and Top-5 accuracy of 5.2\% and 5.7\% by introducing skip connections from the encoder to the decoder.

An example image-sketch pair for each of the four best performing approaches presented in \tblref{tab:results} is given in \figref{fig:user_study}. It shows some of the fundamental qualities of each specific approach. The unconditional approach, denoted as PSim+flip, tends to produce the faintest outlines and fill them with circular inner contours. The conditional approach without any skip connections, denoted as AdaIN, generally outputs the most abstract representations that are often a little bloated and less line-like, but capture the shape of objects well. The two approaches with skip connections are very similar in quality and differ from the prior approaches in their ability to best incorporate finer structures and inner contours.

\subsection{Qualitative Evaluation}\label{sec:quality}

We present further positive and negative reconstruction examples for qualitative evaluation in \figref{fig:example_good_bad}. Our methods work well for objects that appear in a common pose and are framed in an isolated manner, such that there are no other objects immediately adjacent to the object of interest. The top row of \figref{fig:example_good_bad} presents such cases of valid and visually pleasing sketches. Notably, the sketches are not just refined edge images, but rather abstract (and thus human-like) sketches. To emphasize the difference between refined edge maps, and abstract sketches we compare one of the generated sketches (top row of \figref{fig:example_good_bad}, third from the right) with the corresponding (masked) edge map, the object boundary outline and a ground truth example in \figref{fig:edges_mask}. It shows that the (masked) edge map alone is not a useful sketch. The outline of the manually annotated object mask sufficiently conveys the information of the image, but is also exactly what we want to avoid since it is not abstract in nature and does not conform to the human sketching process. The sketch generated with our approach is a more abstract representation that conveys the information without entirely deviating from the input image.

Generally, our method is able to consistently produce sketch-like output images, but sometimes the correspondence with the input image or even the class is lost, despite conditioning of the decoder. Such negative examples are shown in the bottom row of \figref{fig:example_good_bad}. One of the main reasons for reconstruction failures are unusual poses of objects, which is an inherent problem of the sketching process in general. Take as an example the deer in the middle of the row: Here, the decoder assumes a deer, but is unaware of the recumbent pose, and also the sex of the deer. Instead, it generates a deer in profile view, standing and with antlers, which is likely also the way a human sketcher would sketch a deer to convey the underlying concept. For almost any object there appears to exist a preferred, or rather generic pose for sketching, which is also easiest to draw and understand. Any deviation from this generic pose makes the sketching process significantly harder. Other deficiencies include classes that require many strokes to sketch (\eg spider on the left), occlusions (\eg bicycle in the left part), poor visibility of the object (\eg sword on the left), or confusion with adjacent objects (\eg pineapple on the right).

On the whole, classes that are easy to sketch in various poses, such as birds, mugs or kettles, also work best with our method. Objects that are easier to sketch come with more consistent and correct sketches in the dataset and hence cause less confusion in training. See \figref{fig:example_mode} for different images from the classes duck and sailboat. Ducks are easy to sketch in various poses due to characteristic features such as the neck and beak. Although we condition our decoder on the label of the class, the outputs show no sign of mode collapse. A mode collapse is unlikely in our method, because the decoder weights are not optimized with regard to the label. For sailboats, results seemingly indicate a mode collapse, but rather the observed output can be attributed to low variability in the data, especially regarding pose and shape.

\subsection{User study}\label{sec:userstudy}

With generative models, the subjective perceptual image quality of generated samples is hard to quantify. While our proposed method, \ie classifying generated samples with a classifier that is disjoint from the generating method, provides a solid baseline that ensures that the generated samples look sketch-like and belong to the correct class, it is unable to quantify the overall visual quality of a generated sketch. The classifier might still be able to correctly label sketches that are degenerated in a way that human observers would not correctly recognize them.

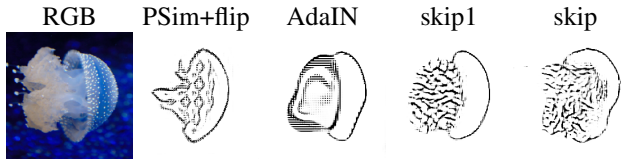
\begin{figure}
\centering
\begin{tikzpicture}[picture format/.style={inner sep=0pt,}]

  \node[picture format] (A1)               {\includegraphics[width=0.2\linewidth]{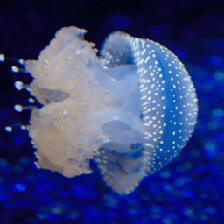}};
  
  \node[picture format,anchor=north west] (A2) at (A1.north east) {\includegraphics[width=0.2\linewidth]{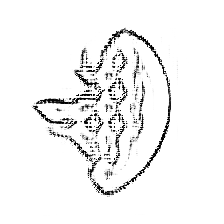}};

  \node[picture format,anchor=north west] (A3) at (A2.north east) {\includegraphics[width=0.2\linewidth]{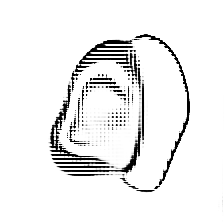}};
  
  \node[picture format,anchor=north west] (A4) at (A3.north east)      {\includegraphics[width=0.2\linewidth]{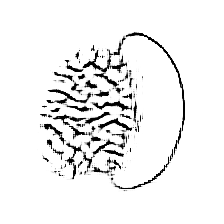}};

  \node[picture format,anchor=north west] (A5) at (A4.north east) {\includegraphics[width=0.2\linewidth]{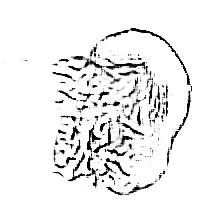}};

  %% Captions

  \node[anchor=base, shift={(0,0.1)}] (C1) at (A1.north) {RGB};
  \node[anchor=base, shift={(0,0.1)}] (C2) at (A2.north) {PSim+flip};
  \node[anchor=base, shift={(0,0.1)}] (C3) at (A3.north) {AdaIN};
  \node[anchor=base, shift={(0,0.1)}] (C4) at (A4.north) {skip1};
  \node[anchor=base, shift={(0,0.1)}] (C5) at (A5.north) {skip};

\end{tikzpicture}
\caption{An example problem set from the user study. We present an RGB image with 4 corresponding sketches from the 4 best performing approaches in \tblref{tab:results} in random order and ask the users to select whichever sketch they like best.}
\label{fig:user_study}
\vspace{-14pt}
\end{figure}

\begin{figure}[t]
\begin{center}
   \includegraphics[width=0.98\linewidth]{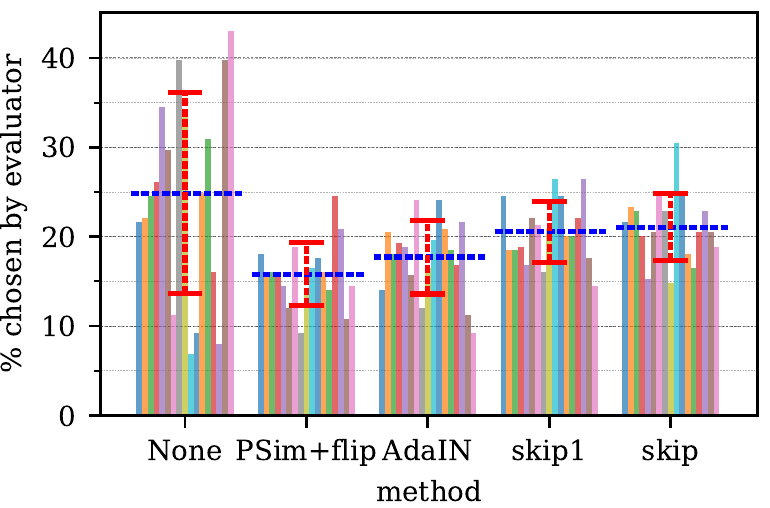}
   \vspace{-4pt}

\end{center}
\vspace{-10pt}
   \caption{Results from the user study. For each method the percentage with which each evaluator has chosen the respective method is shown, with the blue horizontal lines indicating the means and the red vertical bars indicating the standard deviation. The results back up the quantitative evaluation from \tblref{tab:results}: The mean of unclassifiable sketches corresponds well with the Top-1/5 accuracies and the ranking of the individual methods is the same for both quantitative and qualitative evaluation.}
\label{fig:user_study_results}
\vspace{-6pt}
\end{figure}

Hence, we conduct a user study in which we instruct 17 participants (11 male, 6 female, ages 22-61 from various fields of expertise) to compare the four top-performing approaches from \tblref{tab:results} in terms of visual quality. The study consists of 249 trials, 2 sketch-image pairs per class (except for car (sedan) due to dataset limitations), randomly sampled from the test data. Each trial is made up of the RGB image and four sketches from the different methods in random order in the style of \figref{fig:user_study}, but omitting the method labels. On average, the participants took around 25 minutes to complete all of the trials.
\vspace{0pt}
An evaluation of the results is given in \figref{fig:user_study_results}, where the percentage with which each evaluator has selected a given method as well as the per method means and standard deviations are shown. On average, the users decide that 25\% of the sketches are invalid or not classifiable, which falls between the Top-1 and Top-5 accuracies of our quantitatively top performing method. The spread for unclassifiable sketches is rather large compared to the spreads of the different methods, which is in line with expectation, since the decision whether a sketch is valid or not is the most subjective and depends on visual cognition as well as creativity of the beholder. The ranking of the individual approaches corresponds well with the quantitative evaluation in \tblref{tab:results}, although the gap between the \textit{skip1} and \textit{skip} methods is smaller in the qualitative evaluation. In most cases, the visual quality of the two approaches using skip connections is very similar, but for some classes \textit{skip1} does not produce any valid output. It appears that this specific approach is more sensitive to the initialization of the class related mean and standard deviation embeddings. If we were to combine the two different skip approaches into one, there would be a clear preference for the visual quality of approaches that utilize skip connections.

%------------------------------------------------------------------------
%%%%%%%%% CONCLUSION
\section{Conclusion}

We have presented the first approach towards human-like sketch synthesis from natural images via a fully convolutional encoder-decoder structure with a perceptual loss that is able to close this large domain gap. Additionally, we introduced a method that can condition a convolutional decoder on a class prior to further aid the process and show that CNNs are, to the extent of abstracting sketches, able to implicitly represent shape. Our work shows that CNNs trained for image classification learn implicit shape representations and that perceptual loss functions are powerful objectives for image translation tasks between very dissimilar domains. Future work may seek to increase perceptual quality of the generated samples as well as robustness of the method towards odd poses and poorly recognizable images.

{\small
\bibliographystyle{ieee}
\bibliography{bibliography}

\begin{thebibliography}{10}\itemsep=-1pt

\bibitem{chen2018}
W.~Chen and J.~Hays.
\newblock Sketchy{GAN}: Towards diverse and realistic sketch to image
  synthesis.
\newblock In {\em CVPR}, pages 9416--9425, 2018.

\bibitem{dosovitskiy2016}
A.~Dosovitskiy and T.~Brox.
\newblock Generating images with perceptual similarity metrics based on deep
  networks.
\newblock In {\em NeurIPS}, pages 658--666, 2016.

\bibitem{eitz2012}
M.~Eitz, J.~Hays, and M.~Alexa.
\newblock How do humans sketch objects?
\newblock {\em SIGGRAPH}, 31(4):44:1--44:10, 2012.

\bibitem{geirhos2019}
R.~Geirhos, P.~Rubisch, C.~Michaelis, M.~Bethge, F.~A. Wichmann, and
  W.~Brendel.
\newblock Imagenet-trained cnns are biased towards texture; increasing shape
  bias improves accuracy and robustness.
\newblock {\em ICLR}, 2019.

\bibitem{godard2017}
C.~Godard, O.~Mac~Aodha, and G.~J. Brostow.
\newblock Unsupervised monocular depth estimation with left-right consistency.
\newblock In {\em CVPR}, pages 270--279, 2017.

\bibitem{goodfellow2014}
I.~Goodfellow, J.~Pouget-Abadie, M.~Mirza, B.~Xu, D.~Warde-Farley, S.~Ozair,
  A.~Courville, and Y.~Bengio.
\newblock Generative adversarial nets.
\newblock In {\em NeurIPS}, pages 2672--2680, 2014.

\bibitem{ha2018}
D.~Ha and D.~Eck.
\newblock A neural representation of sketch drawings.
\newblock In {\em ICLR}, 2018.

\bibitem{he2015}
K.~He, X.~Zhang, S.~Ren, and J.~Sun.
\newblock Delving deep into rectifiers: Surpassing human-level performance on
  imagenet classification.
\newblock In {\em ICCV}, pages 1026--1034, 2015.

\bibitem{hirata1992}
K.~Hirata and T.~Kato.
\newblock Query by visual example - content based image retrieval.
\newblock In {\em EDBT}, EDBT '92, pages 56--71, London, UK, UK, 1992.
  Springer-Verlag.

\bibitem{huang2017}
X.~Huang and S.~Belongie.
\newblock Arbitrary style transfer in real-time with adaptive instance
  normalization.
\newblock In {\em ICCV}, Oct 2017.

\bibitem{ilg2017}
E.~Ilg, N.~Mayer, T.~Saikia, M.~Keuper, A.~Dosovitskiy, and T.~Brox.
\newblock Flownet 2.0: Evolution of optical flow estimation with deep networks.
\newblock In {\em CVPR}, volume~2, page~6, 2017.

\bibitem{isola2017}
P.~Isola, J.-Y. Zhu, T.~Zhou, and A.~A. Efros.
\newblock Image-to-image translation with conditional adversarial networks.
\newblock {\em CVPR}, pages 5967--5976, 2017.

\bibitem{karras2018}
T.~Karras, S.~Laine, and T.~Aila.
\newblock A style-based generator architecture for generative adversarial
  networks.
\newblock {\em arXiv preprint arXiv:1812.04948}, 2018.

\bibitem{kingma2014}
D.~P. Kingma and J.~Ba.
\newblock Adam: A method for stochastic optimization.
\newblock {\em CoRR}, 2014.

\bibitem{krizhevsky2012}
A.~Krizhevsky, I.~Sutskever, and G.~E. Hinton.
\newblock Imagenet classification with deep convolutional neural networks.
\newblock In {\em NeurIPS}, pages 1097--1105, 2012.

\bibitem{kubilius2016}
J.~Kubilius, S.~Bracci, and H.~P.~O. de~Beeck.
\newblock Deep neural networks as a computational model for human shape
  sensitivity.
\newblock {\em PLoS computational biology}, 12(4):e1004896, 2016.

\bibitem{larsen2016}
A.~B.~L. Larsen, S.~K. S{\o}nderby, H.~Larochelle, and O.~Winther.
\newblock Autoencoding beyond pixels using a learned similarity metric.
\newblock In {\em ICML}, pages 1558--1566, 2016.

\bibitem{mikolov2013}
T.~Mikolov, K.~Chen, G.~Corrado, and J.~Dean.
\newblock Efficient estimation of word representations in vector space.
\newblock {\em CoRR}, abs/1301.3781, 2013.

\bibitem{mirzaO14}
M.~Mirza and S.~Osindero.
\newblock Conditional generative adversarial nets.
\newblock {\em CoRR}, abs/1411.1784, 2014.

\bibitem{muhammad2018}
U.~R. Muhammad, Y.~Yang, Y.-Z. Song, T.~Xiang, and T.~M. Hospedales.
\newblock Learning deep sketch abstraction.
\newblock {\em CVPR}, pages 8014--8023, 2018.

\bibitem{noh2015}
H.~Noh, S.~Hong, and B.~Han.
\newblock Learning deconvolution network for semantic segmentation.
\newblock In {\em ICCV}, pages 1520--1528, 2015.

\bibitem{ridgeway2017}
K.~Ridgeway, J.~Snell, B.~Roads, R.~S. Zemel, and M.~C. Mozer.
\newblock Learning to generate images with perceptual similarity metrics.
\newblock {\em ICIP}, pages 4277--4281, 2017.

\bibitem{ronneberger2015}
O.~Ronneberger, P.~Fischer, and T.~Brox.
\newblock {U-Net: Convolutional Networks for Biomedical Image Segmentation}.
\newblock In {\em MICCAI}, pages 234--241. Springer, 2015.

\bibitem{russakovsky2015}
O.~Russakovsky, J.~Deng, H.~Su, J.~Krause, S.~Satheesh, S.~Ma, Z.~Huang,
  A.~Karpathy, A.~Khosla, M.~Bernstein, A.~C. Berg, and L.~Fei-Fei.
\newblock {ImageNet Large Scale Visual Recognition Challenge}.
\newblock {\em IJCV}, 115(3):211--252, 2015.

\bibitem{saito2017}
M.~Saito, E.~Matsumoto, and S.~Saito.
\newblock Temporal generative adversarial nets with singular value clipping.
\newblock In {\em ICCV}, pages 2830--2839, 2017.

\bibitem{sangkloy2016}
P.~Sangkloy, N.~Burnell, C.~Ham, and J.~Hays.
\newblock The {Sketchy} {Database}: Learning to {Retrieve} {Badly} {Drawn}
  {Bunnies}.
\newblock {\em SIGGRAPH}, 2016.

\bibitem{simonyan2015}
K.~Simonyan and A.~Zisserman.
\newblock Very deep convolutional networks for large-scale image recognition.
\newblock In {\em ICLR}, 2015.

\bibitem{srivastava2014}
N.~Srivastava, G.~Hinton, A.~Krizhevsky, I.~Sutskever, and R.~Salakhutdinov.
\newblock Dropout: a simple way to prevent neural networks from overfitting.
\newblock {\em JMLR}, 15(1):1929--1958, 2014.

\bibitem{szegedy2015}
C.~Szegedy, W.~Liu, Y.~Jia, P.~Sermanet, S.~Reed, D.~Anguelov, D.~Erhan,
  V.~Vanhoucke, and A.~Rabinovich.
\newblock Going deeper with convolutions.
\newblock In {\em CVPR}, pages 1--9, 2015.

\bibitem{szegedy2016}
C.~Szegedy, V.~Vanhoucke, S.~Ioffe, J.~Shlens, and Z.~Wojna.
\newblock Rethinking the inception architecture for computer vision.
\newblock In {\em CVPR}, pages 2818--2826, 2016.

\bibitem{wang2004}
Z.~Wang, A.~Bovik, H.~Rahim~Sheikh, and E.~Simoncelli.
\newblock Image quality assessment: from error visibility to structural
  similarity.
\newblock {\em IEEE transactions on image processing}, 13(4):600--612, 2004.

\bibitem{winnemoller2012}
H.~Winnem{\"o}Ller, J.~E. Kyprianidis, and S.~C. Olsen.
\newblock Xdog: an extended difference-of-gaussians compendium including
  advanced image stylization.
\newblock {\em Computers \& Graphics}, 36(6):740--753, 2012.

\bibitem{xie2015}
S.~Xie and Z.~Tu.
\newblock Holistically-nested edge detection.
\newblock In {\em ICCV}, pages 1395--1403, 2015.

\bibitem{yu2018}
J.~Yu, Z.~Lin, J.~Yang, X.~Shen, X.~Lu, and T.~S. Huang.
\newblock Generative image inpainting with contextual attention.
\newblock In {\em CVPR}, 2018.

\bibitem{yu2017}
Q.~Yu, Y.~Yang, F.~Liu, Y.-Z. Song, T.~Xiang, and T.~M. Hospedales.
\newblock {Sketch-a-Net}: A deep neural network that beats humans.
\newblock {\em IJCV}, 122(3):411--425, May 2017.

\bibitem{zhang2018}
R.~Zhang, P.~Isola, A.~A. Efros, E.~Shechtman, and O.~Wang.
\newblock The unreasonable effectiveness of deep features as a perceptual
  metric.
\newblock In {\em CVPR}, 2018.

\bibitem{zhou2017}
T.~Zhou, M.~Brown, N.~Snavely, and D.~G. Lowe.
\newblock Unsupervised learning of depth and ego-motion from video.
\newblock In {\em CVPR}, pages 5667--5675, 2017.

\end{thebibliography}
}

\end{document}